\pdfoutput=1
\documentclass[11pt]{article}

\usepackage{EMNLP2023}

\usepackage{times}
\usepackage{latexsym}

\usepackage[T1]{fontenc}
\usepackage[utf8]{inputenc}

\usepackage{microtype}

\usepackage{inconsolata}

\usepackage{booktabs}
\usepackage{multirow}
\usepackage{float}
\usepackage{graphicx}
\usepackage{amsmath}
\usepackage{amsfonts}
\usepackage{url}
\usepackage{bm}
\usepackage{xspace}

\usepackage{fdsymbol}

\usepackage{xcolor}

\usepackage{cleveref}

\usepackage{pifont}
\usepackage{color, colortbl}
\definecolor{LightCyan}{rgb}{0.88,1,1}

\title{Learning to Filter Context for Retrieval-Augmented Generation
}

\author{Zhiruo Wang$^{\spadesuit}$ \quad Jun Araki$^{\vardiamondsuit}$ \quad {Zhengbao Jiang}$^{\spadesuit}$ \\
{\bf  Md Rizwan Parvez$^{\vardiamondsuit}$} \quad {\bf Graham Neubig}$^{\spadesuit}$ \\
$^{\spadesuit}$Carnegie Mellon University \quad $^{\vardiamondsuit}$Bosch Research \\
\texttt{\{zhiruow,zhengbaj,gneubig\}@cs.cmu.edu}}

\begin{document}
\maketitle

\begin{abstract}
On-the-fly retrieval of relevant knowledge has proven an essential element of reliable systems for tasks such as open-domain question answering and fact verification. However, because retrieval systems are not perfect, generation models are required to generate outputs given partially or entirely irrelevant passages. This can cause over- or under-reliance on context, and result in problems in the generated output such as hallucinations. 
To alleviate these problems, we propose \textsc{FilCo}, a method that improves the quality of the context provided to the generator by (1) identifying useful context based on lexical and information-theoretic approaches, and (2) training context filtering models that can filter retrieved contexts at test time. 
We experiment on six knowledge-intensive tasks with \textsc{Flan-T5} and \textsc{Llama2}, and demonstrate that our method outperforms existing approaches on extractive question answering (QA), complex multi-hop and long-form QA, fact verification, and dialog generation tasks.
\textsc{FilCo} effectively improves the quality of context, whether or not it supports the canonical output.\footnote{\url{https://github.com/zorazrw/filco}}
\end{abstract}

% main sections
\section{Introduction}
\label{sec:into}

Retrieval augmented approaches to generation have been shown effective for many knowledge-intensive language tasks such as open-domain question answering and fact verification, producing more faithful \citep{khandelwal2020generalization,lewis2020retrieval,shuster-etal-2021-retrieval-augmentation,komeili-etal-2022-internet}, interpretable \citep{guu2020realm}, and generalizable \citep{khandelwal2021nearest} outputs. 
% retrieval is noisy
While the de facto approach is to provide the top retrieved passages to the generator indiscriminately, imperfect retrieval systems often return irrelevant or distracting content.
Generation models are then trained to produce canonical outputs with the guidance of partially or entirely irrelevant passages, and thus are prone to hallucination or spurious memorization.

\begin{figure}[t]
\vspace{-2mm}
    \centering
    \includegraphics[width=0.49\textwidth]{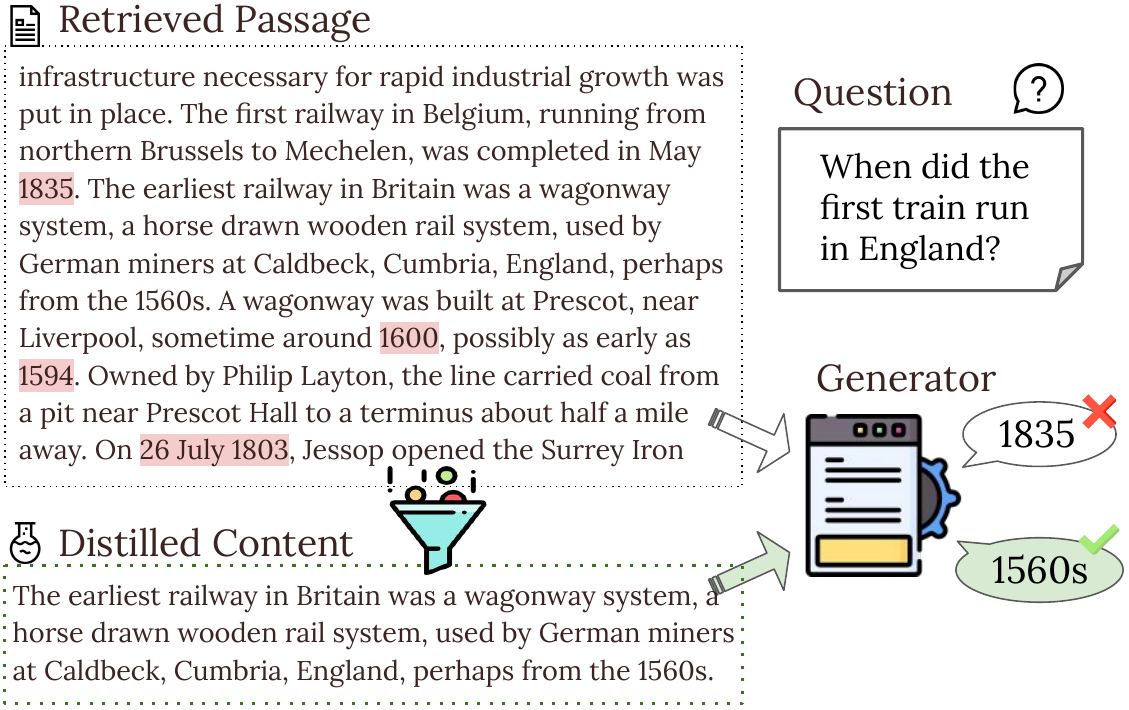}
    \caption{\textsc{FilCo} filters out irrelevant content (marked in red) and leaves precisely supporting content, making it easier for the generator to predict the correct answer.}
\vspace{-2mm}
\label{fig:intro}
\end{figure}

\begin{figure*}[ht]
    \centering
    \includegraphics[width=0.96\textwidth]{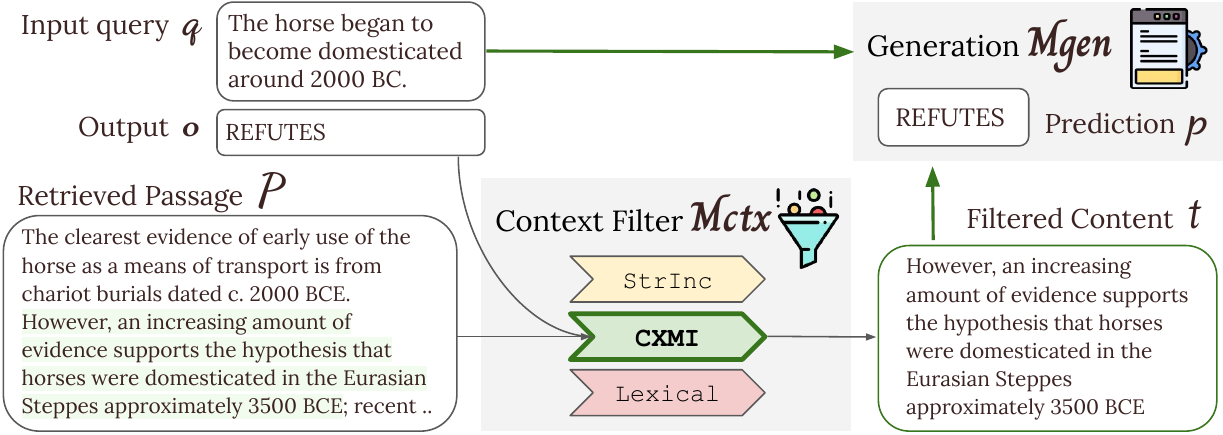}
    \caption{The \textsc{FilCo} pipeline: (i) filtering retrieved passages, (ii) generation with filtered context.}
\label{fig:pipeline}
\end{figure*}

Ideally, a model should be grounded on the precisely supporting content to generate the correct output. However, this ideal grounding is hard to achieve with an imperfect retrieval system alone. 
On one hand, positive passages (i.e., passages that support the output) sometimes contain distracting content. For example in \autoref{fig:intro}, while the passage containing the actual supporting content is successfully retrieved, the model still fails to pay sufficient attention to the supporting content, and is distracted by surrounding sentences that share similar topics \citep{shi2023large}. 
On the other hand, models learn to over-utilize negative passages in the same way as using positive passages, e.g., extracting a span from the irrelevant passage, which would inevitably be incorrect. This potentially degrades accuracy, as training with higher-quality context often leads to better performance \citep{dou-etal-2021-gsum}. 

Some works have attempted to optimize the provided content on the passage level, by reranking more relevant passages rise to the top of the retrieved list \citep{wang2018r,nogueira2020passage,mao-etal-2021-reader}, selecting only evidential passages to include \citep{asai-etal-2022-evidentiality}, or only retrieving passages when generation models need assistance \citep{mallen2023trust,jiang2023active}. 
\citet{choi-etal-2021-decontextualization} proposed to decontextualize sentences by integrating surrounding context, but require substantial human annotation effort and still possibly suffer from distracting content, even in positive passages. 

In this paper, we propose \textsc{FilCo} (\S\ref{sec:filter-context}), a method that learns to \textsc{Fil}ter \textsc{Co}ntext retrieved in a fine-grained sentence-wise manner, by training on content selected via three measures: (i) \textsc{StrInc}: whether passages contain the generation output, (ii) \textsc{Lexical} overlap: how much unigram overlap the content and output has, and (iii) Conditional cross-mutual information (\textsc{cxmi}): how much more likely the generator is to generate the output when the content is provided. 

We experiment on six knowledge-intensive language datasets from three tasks (\S\ref{sec:kilt-data}). (1) question answering: including NaturalQuestions (NQ) and Trivia QA (TQA), as well as more complex multi-hop HotpotQA and long-form ELI5, (2) fact verification: Fact Extraction and VERificaton (FEVER), and (3) knowledge-grounded dialog generation: Wizard of Wikipedia (WoW). 

Using \textsc{Flan-T5} and \textsc{Llama2} models, our method outperforms both baseline methods, i.e., full-context augmentation and passage-wise filtering, on all six datasets. \textsc{FilCo} also greatly reduces the prompt length by $44 - 64 \%$ across tasks.
We further split examples retrieved with positive and negative passages, and show that \textsc{FilCo} effectively improves generation in both scenarios (\S\ref{sec:main-result}). 

Comparing filtering methods on each task, we observe that \textsc{StrInc}, \textsc{Lexical} and \textsc{cxmi}-based filtering were best for extractive QA, dialog generation, and more complex tasks, respectively (\S\ref{sec:filter-ablation}). 
Lastly, we extend experiments to the more complex multi-passage setting, where \textsc{FilCo} maintains its advantage over baseline methods (\S\ref{sec:multi-passage}).

\section{Generation with Filtered Contexts}
\label{sec:filter-context}

In this section, we first outline notation (\S\ref{sub:notation}), then introduce three oracle filtering strategies (\S\ref{sub:oracle-filter}). Next, we describe how to train context filtering models with oracle filtered context (\S\ref{sub:train-filter-model}) and learn to generate with filtered contexts (\S\ref{sub:train-gen-model}).

% ######### %
\subsection{Problem Statement}
\label{sub:notation}

In retrieval-augmented generation, we are given an input query $q$ and annotated output $o$ from an example $e = \{q, o\}$, and want to improve the output of a generative model $M_{gen}$.
We assume a set of retrieved passages $P = \{p_i\}, i \in K$, each consisting of $n_i$ text spans $p_i = [t_i^1, \cdots, t_i^{n_i}]$. 
We can provide the model with one or more selected text spans $T = \{t_i^{j}\}$ when generating output $o$, namely $M_{gen} (o~|~q, T)$. 
In traditional retrieval-based methods, however, all text spans in the top-$K$ passages $\{t_i^{j}\}, \forall j \in n_i, \forall i \in K$ are provided to the model.
In experiments, we split passages into sentences using the spaCy tokenizer\footnote{\url{https://spacy.io/api/tokenizer}} as candidate text spans.
Later in \S\ref{sec:filter-ablation}, we will show that sentence-wise splitting performs the best among other granularities.

% ######### %
\subsection{Obtaining Oracle Contexts}
\label{sub:oracle-filter}

In this section, we propose methods that select oracle text spans that can be used to train a context filtering model. We select spans using a filtering function $F(\cdot)$, denoted as $F(T| e, P)$, where text spans in $T = \{t_i^{j}\}$ are selected by the underlying score function $f(\cdot)$ according to individual filtering methods. We select a single best span $T = t_i^j$, $(i, j) = \operatorname*{arg\,max}_{i,j} f(t_i^j, e)$ when using oracle filtering, as it outperforms multi-span filtering in our preliminary studies. % (\S\ref{app:oracle-text-num}). 

We now introduce three approaches to filtering potentially useful content from retrieved passages.

\paragraph{String Inclusion}
The \textsc{StrInc} measure $f_{inc}(t, o) \in \{0, 1\}$ that makes a binary decision on whether text span $t$ lexically contains the output $o$. We enumerate the ranked passages retrieved $\{p_1, p_2, \cdots \}$ and select the first text span that contains the output $f_{inc}(t, o) = 1$. 
This measure is effective when the supporting document $p_{gold}$ contains the exact output text $o$. However, $f_{inc}$ may fail to distinguish supporting context from spurious ones, that accidentally contain the output but do not answer the question. 
Applying $f_{inc}$ to other abstractive tasks may result in selecting zero spans since no exact matches exist.

\paragraph{Lexical Overlap}
We next introduce a more flexible \textsc{Lexical} measure $f_{uf1} \in [0, 1]$ that calculates the unigram overlap between the example $e$ and the candidate text span $t$. Intuitively speaking, higher lexical overlap indicates greater topic similarity, hence higher utility at generation time.

We select sentences $t$ using different parts of the example $e$ for tasks of different types. We measure the F$_1$ score $f_{uf1}(t, o) \in [0, 1]$ between $t$ and output $o$ for tasks having responses grounded on provided knowledge, i.e., QA and dialog generation. We measure $t$ using query $q$ for fact verification as $f_{uf1}(t, q)$ since $o$ is a one-word binary label. 
We select the sentence $t_i^j$ with the highest similarity to example $e$ and above a pre-defined threshold $\lambda = 0.5$,\footnote{We compare different thresholds ($0.1$, $0.3$, $0.5$, $0.7$, $0.9$) in preliminary studies, $0.5$ gives the best generation results.} where $(i, j) = \operatorname*{arg\,max}_{i,j} (f_{uf1}(t_i^j, e))$, and $i, j \in \{i, j~|~f_{uf1}(t_i^j, e) > \lambda \}$. 
Nonetheless, for tasks having queries that may be factually incorrect (e.g., fact verification), spans of high lexical overlap to an erroneous claim may reinforce the misinformation and lead to incorrect generations.

\paragraph{Conditional Cross-Mutual Information (CXMI)}

We adopt a measure $f_{cxmi}$ from the conditional cross-mutual information (CXMI) score in contextual machine translation \cite{fernandes-etal-2021-measuring}. 

\begin{figure}[ht]
\centering
    \includegraphics[width=0.50\textwidth]{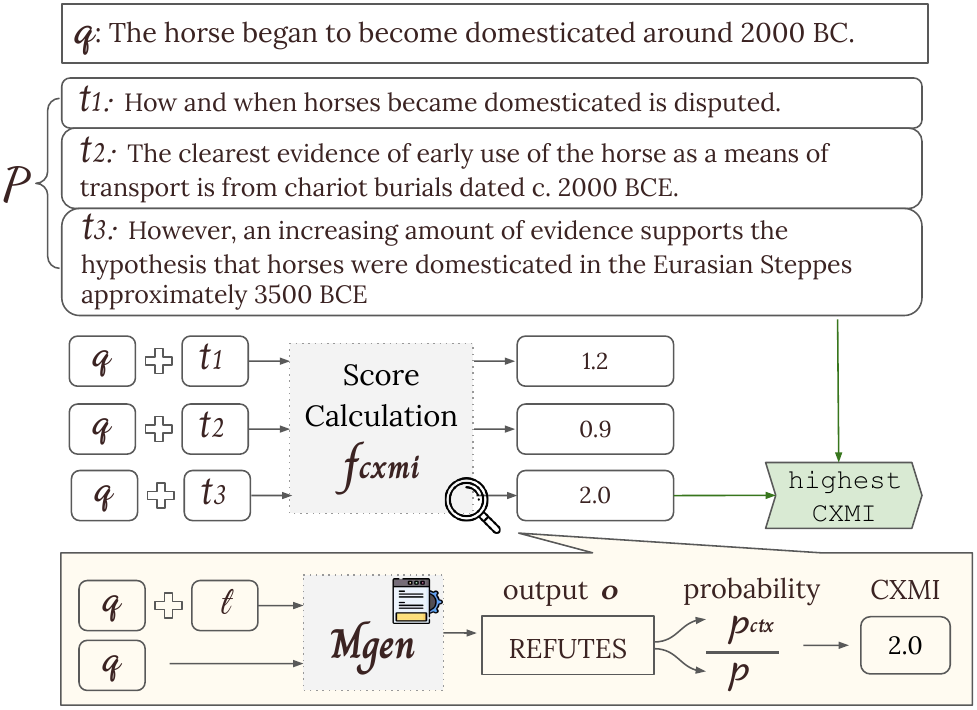}
    \caption{An example illustration of context filtering with the \textsc{cxmi} strategy. }
\vspace{-1mm}
\label{fig:silver-filter}
\end{figure}

Given a pair of input sequences with and without context augmentation, $t \oplus q$ and $q$, we measure the probability difference in model $M_{gen}$ generating the expected output $o$, the process being denoted as $f_{cxmi}(t, e) = \frac{M_{gen}(o | t \oplus q)}{M_{gen}(o | q)} \in \mathbb{R}$, as illustrated in \autoref{fig:silver-filter}. 
% Because \textsc{cxmi} scores can be infinitely large in magnitude, we normalize the scores to $[0, 1]$ by \textit{sigmoid} to ease human and machine understanding. 
We select the text span $t_i^j$ having the highest \textsc{cxmi} score above a pre-defined threshold $\lambda = 0.0$,\footnote{$1.0$ naturally distinguishes context that adds to or reduces output probability. We compare other values in preliminary studies ($0.5$, $2.0$), where $1.0$ gives the best results.} where $(i, j) = \operatorname*{arg\,max}_{i,j} (f_{cxmi}(t_i^j, e))$, and $i, j \in \{i, j~|~f_{cxmi}(t_i^j, e) > \lambda \}$. 
$f_{cxmi}$ can overcome the lexical barrier and is applicable to all tasks, however at the cost of more computation.

% ######### %
\subsection{Learning to Filter Contexts}
\label{sub:train-filter-model}
While the previous section described how to identify useful contexts at training time when the gold standard answer is known, we also need methods that can apply at test time when the answer is unknown. To this end, we train the context filtering models, $M_{ctx}$, using context filtered with the three measures in \S\ref{sub:oracle-filter}.
To create training data for $M_{ctx}$, for each training example with query $q$, we concatenate the retrieved passages $P$ and query $q$ as input, then, we apply the filter method $f$ to obtain filtered context $t_{silver}$ as output. We use ${silver}$ instead of ${oracle}$ to represent the non-perfect filtering result due to unknown gold labels for non-extractive tasks. 
As shown in \autoref{fig:pipeline}, we train $M_{ctx}$ by feeding in query $q$ and retrieved passages $P$, and ask it to generate filtered context $t_{silver}$, formalized as $M_{ctx}(t_{silver} | ~q \oplus P)$.

At test time, given the retrieved passages $P$ for each test query $q$, we leverage $M_{ctx}$ to predict filtered context $t_{pred}$, formalized as $t_{pred} = M_{ctx}(q \oplus P)$. $t_{pred}$ is subsequently provided to the generation model $M_{gen}$ together with the query $q$, to predict the output.

% ######### %
\subsection{Generation With Filtered Contexts}
\label{sub:train-gen-model}
As illustrated in \autoref{fig:pipeline}, we similarly use $t_{silver}$ filtered context for training and model predicted context $t_{pred}$ for inference. 

For each training example $(q, o)$, we prepend the silver filtered context $t_{silver}$ to the example query $q$, and obtain the model input $q \oplus t_{silver}$. We feed this input into the generation model $M_{gen}$ and train it to output the canonical response $o$, formalized as $M_{gen}(o~|~ t_{silver} \oplus q)$. 

At inference time, we provide the context $t_{pred}$ filtered by model $M_{ctx}$ for generation, denoted as $M_{gen}(o~| ~t_{pred} \oplus q) = M_{gen}(o~|~M_{ctx}(q, P) \oplus q)$. 

In comparison to appending all retrieved text spans $P \oplus q$, including only selected text can effectively reduce the computational cost by $\frac{|P|}{|t|}$ at both training and inference time.

\section{Knowledge-Intensive Language Tasks}
\label{sec:kilt-data}

We experiment on six knowledge-intensive language tasks that necessitate retrieval augmentation for generation (\S\ref{sub:task-and-data}), where a limited portion of examples are supported by retrieved passages (\S\ref{sub:retrieval-result}).

% ################### %
\subsection{Tasks and Datasets}
\label{sub:task-and-data}

We use six datasets built from Wikipedia articles as supporting documents for answer, response, and judgment generation, as listed in \autoref{tab:dataset-stats}.

\paragraph{Open-Domain Question Answering} 
We adopt NaturalQuestions (NQ) \cite{kwiatkowski-etal-2019-natural} and TriviaQA (TQA) \cite{joshi-etal-2017-triviaqa} to experiment with the open-domain QA task. 

Each example in NQ has a question $q$ and annotated short answers $o$. We experiment with the processed version \citep{lee-etal-2019-latent} that includes all examples having short answers of no more than five tokens. 
For the TQA dataset, each example has a question $q$ and answers $o$, which are extracted spans from supporting Wikipedia articles $P$. 
Following \citet{lewis2020retrieval}, we use the Exact Match (EM) metric to evaluate model predictions.

\paragraph{Multi-Hop Question Answering}
We also adopt more complex QA scenarios, the first of which is multi-hop QA, where each question $q$ requires reasoning over a chain of passages $P$ to obtain the correct answer $o$. 
For this task, we use the HotpotQA \citep{yang-etal-2018-hotpotqa} dataset containing 113$K$ question-answer pairs created based on Wikipedia pages. Because the answers $o$ do not always appear in the ground-truth supporting documents $P$, this dataset belongs to abstractive generation, in contrast to the extractive nature of answers in NQ and TQA.
Following \citet{yang-etal-2018-hotpotqa} and accommodating its abstractive nature, we use unigram F$_1$ to evaluate answer correctness.

\paragraph{Long-Form Question Answering}
Another complex QA task is generating long, abstract answers given the question, i.e., long-form QA. For this we use the ELI5 \citep{fan-etal-2019-eli5} dataset, which requires elaborate and in-depth answers to open-ended questions. The dataset comprises 270K threads from the Reddit forum ``Explain Like I’m Five'' (ELI5) and features diverse questions requiring multi-sentence answers. We experiment with the \emph{generative short} setting, and evaluate model predictions using unigram F$_1$.

\paragraph{Fact Verification}
We use the Fact Extraction and VERification (FEVER) dataset \cite{thorne-etal-2018-fever} aggregated by the KILT benchmark \citep{petroni-etal-2021-kilt}. It contains claims $q$ generated by rephrasing sentences in Wikipedia articles. A claim has the label $o=$ ``SUPPORTS'' if it preserves the fact in the Wikipedia reference, otherwise is labeled as ``REFUTES'' due to the fact contradiction.
Following the original baseline \citep{thorne-etal-2018-fever}, we use accuracy for evaluation.

\paragraph{Knowledge-Grounded Dialog Generation}
We adopt the Wizard of Wikipedia (WoW) dataset \cite{dinan2018wizard} from KILT, which aims to generate the next dialog by grounding on Wikipedia articles. In each example, the input $q$ is the conversation history involving multiple utterance turns, and the next-turn response is the output $o$. We evaluate with unigram F$_1$ following \citet{petroni-etal-2021-kilt}.

\begin{table}[ht]
\vspace{1mm}
\small 
\centering 
\resizebox{0.49\textwidth}{!}{
    \begin{tabular}{l|ccc|c}
    \toprule
    \multirow{2}{*}{\textbf{Dataset}} & \multicolumn{3}{c|}{\textbf{\# Examples} (thousands)} & {\textbf{Evaluation}} \\
    {} & {\textbf{train}} & \textbf{dev} & \textbf{test} & {\textbf{Metric}} \\
    \midrule 
    \textsc{NQ} & {$~~$79.2} & {$~~$8.7} & {$~~$3.6} & {EM} \\
    \textsc{TQA} & {$~~$78.8} & {$~~$8.8} & {11.3} & {EM} \\
    % \midrule
    \textsc{HotpotQA} & {$~~$88.9} & {$~~$5.6} & {$~~$5.6} & {F$_1$} \\
    \textsc{ELI5} & {273.0} & {$~~$1.5} & {$~~$0.6} & {F$_1$} \\
    % \midrule
    \textsc{FEVER} & {105.0} & {10.4} & {10.1} & {Accuracy} \\
    % \midrule
    \textsc{WoW} & {$~~$63.7} & {$~~$3.1} & {$~~$2.9} & {F$_1$} \\
    \bottomrule
    \end{tabular}
}
\caption{Statistics and evaluation metric for six tasks.}
\vspace{1mm}
\label{tab:dataset-stats}
\end{table}

\autoref{tab:dataset-stats} lists the dataset statistics. Because test sets are not available for datasets adopted from the KILT benchmark (i.e., HotpotQA, ELI5, FEVER, WoW), we report the development set results.

% ################### %
\subsection{Wikipedia Passage Retrieval}
\label{sub:retrieval-result}

To better understand the quality of passages provided in the generation stage, we evaluate the performance of retrieval results. 

To retrieve Wikipedia passages for all examples, we use the adversarial Dense Passage Retriever (DPR) \citep{karpukhin-etal-2020-dense}\footnote{\url{https://github.com/facebookresearch/DPR\#new-march-2021-retrieval-model}} to retrieve the top 5 passages from all Wikipedia passages.

\paragraph{A Mixture of Positive and Negative Passages}
We evaluate the \textit{recall} of the top 1 and top 5 retrieved passages in \autoref{tab:retrieval-results}. For the extractive NQ and TQA tasks, we measure if any of the passages contain the answer strings. For the other four tasks where outputs are not spans in supporting documents, we calculate if any of the passages come from the provenance articles annotated in KILT.

Notably, for all six datasets, top-1 passages only support the canonical output half or less of the time. Although involving more passages increases the coverage of supporting documents, it often brings along linearly \citep{izacard-grave-2021-leveraging} or quadratically increased computation.

\begin{table}[ht]
% \vspace{-1mm}
\small 
\centering 
\resizebox{0.49\textwidth}{!}{
    \begin{tabular}{l|cc|cc}
    \toprule
    \multirow{2}{*}{\textbf{Dataset}} & \multicolumn{2}{c|}{\textbf{Recall} (pos. + neg.)} & \multicolumn{2}{c}{\textbf{Precision} (pos.)} \\
    {} & \textbf{1} & \textbf{5} & \textbf{1} & \textbf{5} \\
    \midrule 
    \textsc{NQ} & {50.1} & {74.1} & {$~~$2.5} & {$~~$2.7} \\
    \textsc{TQA} & {61.2} & {77.8} & {$~~$4.5} & {$~~$4.8} \\
    \textsc{HotpotQA} & {16.7} & {27.3} & {$~~$2.1} & {$~~$0.4} \\
    \textsc{ELI5} & {13.1} & {25.7} & {97.7} & {55.1} \\
    \textsc{FEVER} & {57.0} & {75.9} & {$~~$1.3} & {$~~$1.4} \\
    \textsc{WoW} & {34.9} & {54.8} & {16.4} & {17.7} \\
    \bottomrule
    \end{tabular}
}
\caption{Recall of the top 1 and top 5 DPR-retrieved passages, and precision on positive passages.}
\vspace{-2mm}
\label{tab:retrieval-results}
\end{table}

\paragraph{Noise in Positive Passagess}
To measure the ratio of precisely supporting context in retrieved passages, we further calculate their unigram \textit{precision} with regard to the annotated output, as shown in \autoref{tab:retrieval-results}. 
In general, the precision is pretty low: scoring less than $20.0$ for WoW, and less than $5.0$ for NQ, TQA, HotpotQA, and FEVER. ELI5 has exceptionally high top-1 precision, because its output often aggregates large text chunks from multiple passages. However, precision drops by over 40 points when adding 4 more passages.
These numbers indicate the potential existence of redundant content, which could distract the model and deteriorate the final generation. 

In the next section, we attempt to filter the sufficient and precisely necessary context, as described in \S\ref{sec:filter-context}, to achieve more efficient generation.
\section{Experiments and Analysis}
\label{sec:main-result}

We first introduce the experimental setup (\S\ref{sub:impl-detail}) and baseline approaches for comparison (\S\ref{sub:baseline}). Then, we evaluate model performance on both end generation (\S\ref{sub:gen-perf}) and context filtering (\S\ref{sub:filter-perf}).

% ############# %
\subsection{Experimental Setup}
\label{sub:impl-detail}
We use \textsc{Flan-T5} \citep{chung2022scaling} and \textsc{Llama 2} \citep{touvron2023llama} as the backbone model architectures, because of their potential superior performance among open-source models. 
We fine-tune both models for (i) the context filtering task as $M_{ctx}$, and (ii) the end generation task as $M_{gen}$. 

\paragraph{\textsc{Flan-T5}} 
\textsc{Flan-T5} is a family of instruction-tuned encoder-decoder models for seq2seq generation tasks, which makes it suitable for our retrieval-augmented generation setting. Due to constraints in computational resources, we use the \textsc{XL} version with 3$B$ parameters. We load model checkpoints from and implement training using HuggingFace Transformers \citep{wolf-etal-2020-transformers}.

\paragraph{\textsc{Llama 2}}
\textsc{Llama 2} represents a collection of foundation model ranging from 7$B$ to 70$B$ parameters, particularly optimized for dialog uses cases, but also achieve good performance on many other tasks. We train the 7$B$ model version with LoRA \citep{hu2022lora} using the xTuring platform.\footnote{\url{https://github.com/stochasticai/xTuring}}

\paragraph{Implementation Details}
For both models, we allow a maximum length of 1024 tokens for all sequences at training and inference. $M_{ctx}$ is configured to generate at most 512 tokens as filtered context for all tasks. We allow $M_{gen}$ to generate at most 128 tokens for extractive QA, fact verification, and dialog generation tasks. We use greedy decoding for generating both filtered context and end-generation output. Unless otherwise specified, we train all $M_{ctx}$ and $M_{gen}$ models for $3$ epochs, using a learning rate of $5\mathrm{e}{-5}$ and batch size of $32$.

\begin{figure*}[h]
\vspace{-4mm}
    \centering
    \includegraphics[width=\textwidth]{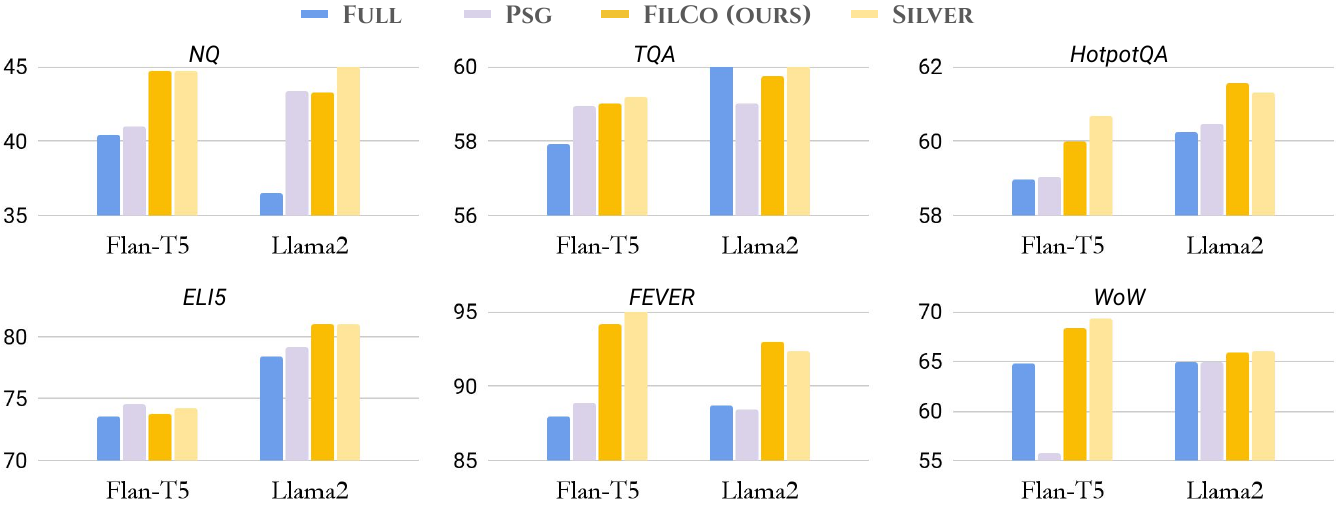}
    \caption{Generation performance when passages are filtered with different approaches.}
\vspace{-1mm}
\label{fig:main-results}
\end{figure*}

% ############# %
\subsection{Experiment Methods}
\label{sub:baseline}

We describe two baselines \textsc{Full} and \textsc{Psg}, our main approach \textsc{FilCo}, and the \textsc{Silver} setting.

\paragraph{Baseline 1: Augmenting with Full Passages}
The most common approach for retrieval-augmented generation is to concatenate all passages into the input. We denote this method as \textsc{Full} and adopt it as our first baseline. To conduct a fair comparison with sufficient training for generation in a full-context style, we fine-tune the \textsc{Flan-T5} and \textsc{Llama 2} models to generate outputs using the full content of the top-1 passages under the same experiment setting as in \S\ref{sub:impl-detail}.

\paragraph{Baseline 2: Passage-Wise Filtering}
An alternative method inspired by \citet{asai-etal-2022-evidentiality} is to filter context on a passage level. Specifically, for each passage among the top-1 retrieved ones, the model decides whether to include the entire piece of the passage in the input. In comparison, our method operates in a finer granularity (i.e., on the sentence level) and could trained with multiple filtering strategies. To show the empirical advantage of our method, we denote this approach as \textsc{Psg} and adopt it as another baseline.

\paragraph{Main Approach: Augmenting with Filtered Context}
As described in \S\ref{sec:filter-context}, we train $M_{ctx}$ to filter the top-1 retrieved passage $P$ to $t_{silver}$, and $M_{gen}$ to generate output $o$ with $t_{silver}$. To create $t_{silver}$, we use the \textsc{StrInc} measure for NQ and TQA, \textsc{Lexical} for FEVER, and \textsc{cxmi} for WoW, HotpotQA, and ELI5. These measures are shown to be the optimal settings based on further analysis in \S\ref{sec:filter-ablation}. 

At test time, we provide model-filtered context $t_{pred}$ to $M_{gen}$, and denote the results as \textsc{FilCo}. 
To demonstrate the prospective performance upper-bound, we also evaluate $M_{gen}$ generation by providing silver-filtered context $t_{silver}$, and denote these results as \textsc{Silver}.

% ############# %
\subsection{Generation Performance}
\label{sub:gen-perf}

Results using four methods and two models are shown in \autoref{fig:main-results}. In general, applying context filtering beforehand significantly improves the results on all datasets than \textsc{Full}. Moreover, filtering in a finer granularity is better than \textsc{Psg}.

Compared to providing $M_{gen}$ with \textsc{Silver} filtered contexts, using contents predicted by the filter model, i.e., \textsc{FilCo} achieves comparable performance on all six tasks, indicating effective training of the context filtering process. 

For \textit{extractive QA tasks}, our method achieves $+ 4.3$ and $+ 8.6$ EM increase in NQ with \textsc{Flan-T5} and \textsc{Llama2} models, $+ 1.1$ and $+ 0.2$ EM increase in TQA. As exemplified by \autoref{fig:intro}, our context filter effectively removes distracting alternative answers and irrelevant passages, hence enabling the generation model to hit the correct answer span with higher precision and lower effort. 

For \textit{more complex QA tasks}, our method brings $+ 1.0$ and $+ 1.3$ F$_1$ increase in HotpotQA with \textsc{Flan-T5} and \textsc{Llama2} models, and $+ 0.6$, $+ 2.6$ EM increase in ELI5. The overall improvement is less significant, compared to extractive QA tasks, presumably due to the increased task difficulty.

For \textit{abstractive generation tasks}, our method brings about even larger improvements: $+ 6.2$ and $+ 4.3$ accuracy increase for FEVER with \textsc{Flan-T5} and \textsc{Llama2}, and $+ 3.5$, $+ 1.1$ F$_1$ increase for WoW. As could be partially conjectured from the low precision in \autoref{tab:retrieval-results}, filtering irrelevant content helps the model focus on the concerned knowledge.

\begin{figure*}[ht]
\vspace{-3mm}
    \centering
    \includegraphics[width=\textwidth]{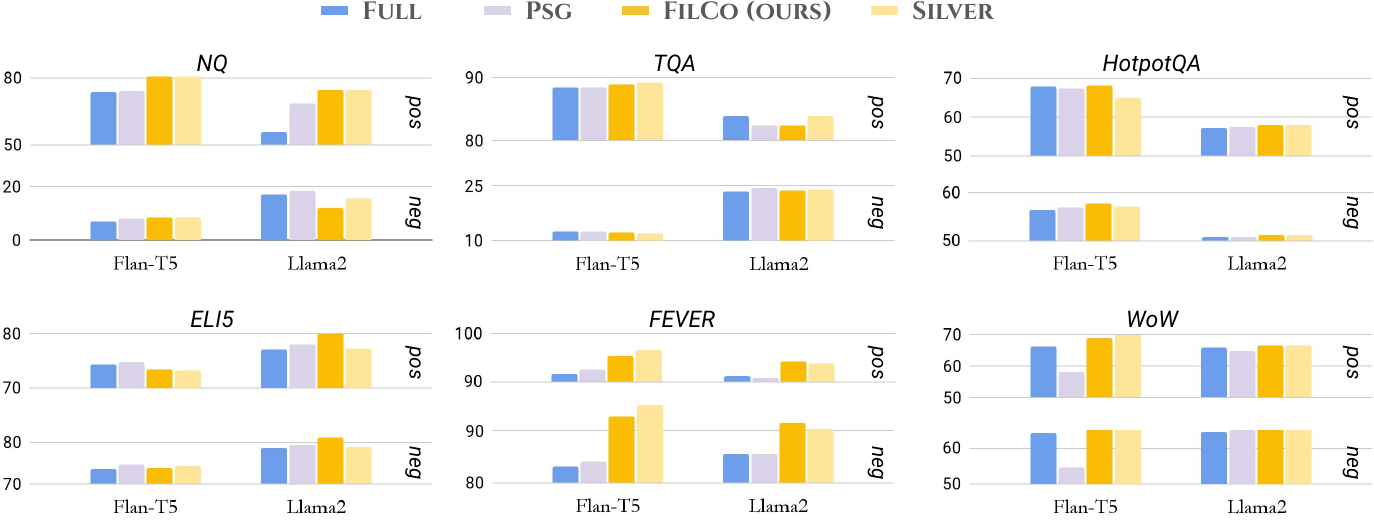}
    \caption{Improvement on examples retrieved with positive (top) and negative passages (bottom), respectively.}
\label{fig:pos-neg}
\end{figure*}

\subsection{Generation With Filtered Positive and Negative Passages}

We decompose datasets into examples with positive and negative top-1 retrieved passages, to examine improvements under both scenarios.

As shown in \autoref{fig:pos-neg}, for both positive and negative passages retrieved, applying \textsc{FilCo} effectively improves the context quality, hence yields better end generation results, particularly for abstractive generation tasks such as FEVER and WoW. 
Aligning with our hypothesis, the generation model produces more correct outputs when we remove (i) distracting content in positive passages, and (ii) negative passages.

% ############### %
\subsection{Evaluating Filtered Contexts}
\label{sub:filter-perf}

We evaluate context filtering outputs from two aspects: reduced input length and increased answer precision.

\begin{figure}[ht]
    \centering
    \includegraphics[width=0.5\textwidth]{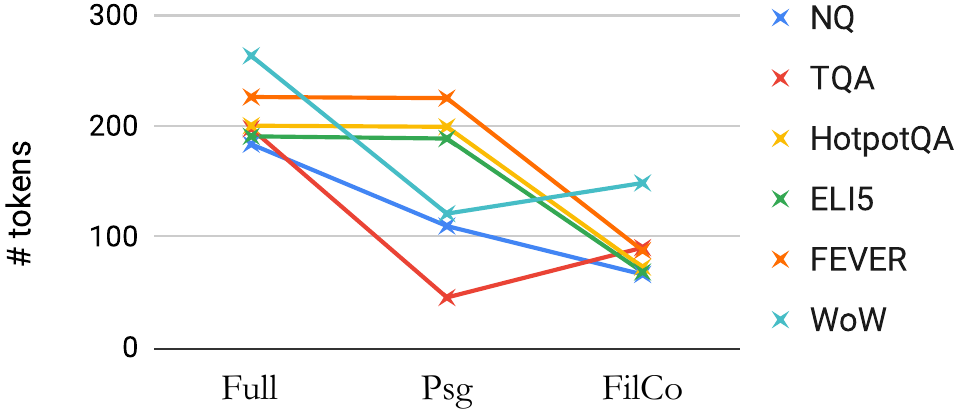}
    \caption{Number of input tokens after filtering retrieved contexts with different strategies.}
    \vspace{-2mm}
\label{fig:num-input-tokens}
\end{figure}

\paragraph{Shorter Inputs}
In \autoref{fig:num-input-tokens}, we measure the average number of tokens in model inputs after filtering the retrieved contexts using different methods. More specifically, we do not filter context in the \textsc{Full} setting, filter context by passage in the \textsc{Psg} setting, and filter context in the sentence level with \textsc{FilCo}.
Model inputs contain the original query and (filtered) context.\footnote{We tokenize all text sequences with the LlamaTokenizer.}
Our method (the \textsc{FilCo} column) effectively reduces input length by $44 - 64 \%$.

\paragraph{Higher Precision}
To evaluate the amount of potentially redundant information in the context, we measure the unigram precision of outputs with respect to filtered or unfiltered contexts.

As shown in \autoref{tab:eval-filter}, context after filtering achieves much higher precision for all tasks. Particularly for abstractive tasks, \textsc{Silver} filtering increases the precision by $+ 14.5$ on HotpotQA and $+ 60.7$ on WoW.
Moreover, model-filtered contexts (\textsc{FilCo}) are largely comparable to \textsc{Silver}, and sometimes even better, such as $+ 3.8$ points in TQA. For other tasks, the small gaps between them minimally affect the end generation, as already shown in \autoref{fig:main-results}. We conjecture these lost contents are not essential for models, particularly if they only involve common entities \citep{mallen2023trust}. 

However, filtering with the \textsc{Psg} baseline often leads to precisions lower than the \textsc{Full} setting, despite the fact that it has higher output scores than \textsc{Full}. Coarse granularity for context filtering may be one major reason for its loss in precision.

\begin{table}[ht]
\small 
\centering
\resizebox{0.47\textwidth}{!}{
    \begin{tabular}{l|cccc}
    \toprule
    \textbf{Method} & \textbf{\textsc{Full}} & \textbf{\textsc{Psg}} & \textbf{\textsc{FilCo}} & \textbf{\textsc{Silver}} \\
    \midrule
    \textsc{NQ} & {$~~$2.5} & {$~~$1.3} & {$~~$5.1} & \textbf{$~~$7.3} \\
    \textsc{TQA} & {$~~$4.5} & {$~~$3.0} & \textbf{$~~$8.4} & {$~~$4.6} \\
    \textsc{HotpotQA} & {$~~$2.6} & {$~~$2.6} & {10.8} & \textbf{17.1} \\
    \textsc{ELI5} & {92.9} & {92.5} & \textbf{98.8} & \textbf{98.8} \\
    \textsc{FEVER} & {$~~$1.2} & {$~~$1.2} & \textbf{$~~$5.1} & {$~~$4.4} \\
    \textsc{WoW} & {10.8} & {35.5} & {62.9} & \textbf{71.5} \\
    \bottomrule
    \end{tabular}
}
\caption{Precision of canonical outputs with respect to contexts filtered with different methods.}
\vspace{-2mm}
\label{tab:eval-filter}
\end{table}

\section{Comparing Context Filtering Strategies}
\label{sec:filter-ablation}

To justify the selection of context filtering strategies in \S\ref{sec:main-result}, we compare different measures to create filter training data, as introduced in \S\ref{sec:filter-context}.

\subsection{Results with Different Strategies}

We compare using three methods in \S\ref{sub:oracle-filter} --- \textsc{StrInc}, \textsc{Lexical}, and \textsc{cxmi} --- to train the context filter model, on datasets of different properties. 

Results in \autoref{tab:silver-selection} reveal that different tasks benefit the most from different measures. NQ and TQA favor \textsc{StrInc}, WoW works best with \textsc{Lexical}, while more complex tasks such as \textsc{FEVER}, \textsc{HotpotQA}, and \textsc{ELI5} perform the best using \textsc{cxmi}.
Model wise, \textsc{Flan-T5} and \textsc{Llama2} align on most tasks, with slight divergence on ELI5.

\begin{table}[ht]
\vspace{-1mm}
\small 
\centering
\resizebox{0.43\textwidth}{!}{
    \begin{tabular}{l|ccc}
        \toprule
        \multicolumn{4}{c}{\textsc{Flan-T5}} \\
        \midrule
        \textbf{Measure} & \textbf{\textsc{StrInc}} & \textbf{\textsc{Lexical}} & \textbf{\textsc{cxmi}} \\
        \midrule
        \textsc{NQ} & \textbf{44.7} & {30.0} & {39.9} \\
        \textsc{TQA} & \textbf{59.2} & {39.0} & {45.3} \\
        \textsc{HotpotQA} & {59.2} & {57.4} & \textbf{60.0} \\
        \textsc{ELI5} & {73.6} & {73.9} & \textbf{74.2} \\
        \textsc{FEVER} & {80.9} & {86.4} & \textbf{95.8} \\
        \textsc{WoW} & {63.4} & \textbf{69.3} & {66.6} \\
        \midrule
        \multicolumn{4}{c}{\textsc{Llama 2}} \\
        \midrule
        \textsc{NQ} & \textbf{43.3} & {35.2} & {41.8} \\
        \textsc{TQA} & \textbf{60.7} & {57.1} & \textbf{60.7} \\
        \textsc{HotpotQA} & {59.5} & {61.1} & \textbf{61.3} \\
        \textsc{ELI5} & {78.6} & \textbf{78.8} & {72.8} \\
        \textsc{FEVER} & {86.6} & {88.4} & \textbf{92.3} \\
        \textsc{WoW} & {65.5} & \textbf{66.0} & {65.4} \\
        \bottomrule
    \end{tabular}
}
\caption{\textsc{Flan-T5} and \textsc{Llama2} using different context filtering measures on each dataset.}
\vspace{-1mm}
\label{tab:silver-selection}
\end{table}

\subsection{In-Depth Analysis for Different Tasks}

\textit{Extractive tasks} (i.e., NQ and TQA) achieve the best with an \textsc{StrInc} context filter. This phenomenon reasonably aligns with their extractive nature, where sentences that lexically entail the output answer usually are the ground truth supporting content, except for the case of spurious contexts.\footnote{Context that accidentally contains the answer string but does not actually answer the question.} 

On the other hand, the \textsc{StrInc} strategy falls short on \textit{abstractive tasks} (i.e., FEVER and WoW), due to the task feature that canonical output does not exist in supporting documents, hence would often yield empty content for subsequent generation. 
An adapted \textsc{Lexical} measure F$_1$ readily allows more flexible unigram-level matches and is the most suitable approach for \textit{dialog generation}. 

\begin{figure}[h]
\vspace{2mm}
    \centering
    \includegraphics[width=0.48\textwidth]{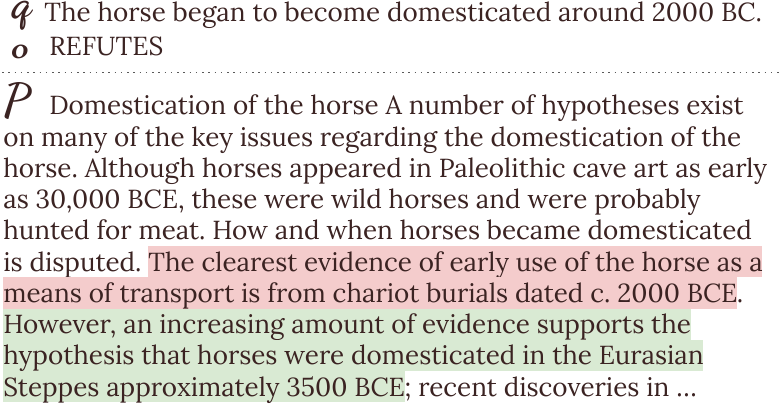}
    \caption{An example in the FEVER dataset illustrating filtering outcomes using different strategies. \textsc{StrInc} yields empty context, \textsc{Lexical} and \textsc{cxmi}-filtered context are highlighted in \colorbox{red!20}{red} and \colorbox{green!20}{green}, respectively.}
\vspace{-1mm}
\label{fig:comp-silver-fever}
\end{figure}

\textsc{cxmi} works the best for more complex tasks, i.e., \textit{multi-hop QA}, \textit{long-form QA}, and \textit{fact verification}.

Taking the fact verification task for example, while factually incorrect claims (labeled as REFUTES) often contain entity spans from irrelevant or distracting content in the passage, such as ``2000 BC'' in \autoref{fig:comp-silver-fever}, lexical measures would falsely pick the \colorbox{red!20}{misleading content} that matches ``2000 BC'' but concerns about ``evidence of early use'' instead of ``become domesticated''. Augmenting with this content can reinforce the spurious correlation via the misleading fact (``2000 BC'') and deteriorate the generation performance. In comparison, selecting only the \colorbox{green!20}{content supportive} of making factual judgment can provide the correct knowledge that horses became domesticated around ``3500 BC''.

\section{Generation with Multiple Passages}
\label{sec:multi-passage}

It is often helpful to integrate multiple passages as context input to the model. Particularly, some tasks such as multi-hop QA may naturally necessitate using multiple passages to perform the task.
To demonstrate the generality of our proposed method, we further experiment using multiple passages as source context.
We experiment with \textsc{Flan-T5} since it has more consistent behaviors across tasks.

\subsection{Baseline and Settings}
We experiment with top-K passages, where $K = 5$, to minimize the loss from length truncation due to model input limitations, compared to larger $K$s, and hence produce more fair comparisons.

Similarly to the single-passage setting, we compare \textsc{Full} and \textsc{Psg} as baseline methods, where \textsc{Full} inputs all passages unfiltered and \textsc{Psg} picks zero or more passages.
We also include the results of top-performing methods such as RAG \citep{lewis2020retrieval}, FiD \citep{izacard-grave-2021-leveraging}, and evidentiality-guided (\textsc{Evi.}) generation \citep{asai-etal-2022-evidentiality}. 
In comparison to baselines, we report the sentence-wise filtering method as \textsc{FilCo} and the canonical setting by \textsc{Silver}.

\subsection{Generation Performance}

As shown in \autoref{tab:top5-granularity}, our main method \textsc{FilCo} surpasses the full-context (\textsc{Full}) and passage-filtering (\textsc{Psgs}) settings by a large margin, $+ 1.2 - 14.2$ points in all six tasks. \textsc{FilCo} also outperforms existing performant baselines. 
Compared to using top-1 passages only, performance increases on extractive tasks when aggregating multiple top-ranked passages. Interestingly, performance on FEVER and WoW drop by $-3.2$ and $-2.3$ points, potentially due to the decreased retrieval quality of lower-ranked passages, as the top-1 retrieval recall is relatively high. 

\begin{table}[ht]
\small
\centering
\resizebox{0.49\textwidth}{!}{
    \begin{tabular}{l|cccccc}
    \toprule
    \textbf{Context} & \textbf{NQ} & \textbf{TQA} & \textbf{HotpotQA} & \textbf{ELI5} & \textbf{FEVER} & \textbf{WoW} \\
    \midrule
    \multicolumn{7}{c}{\textsc{Baseline, top 5}} \\
    \midrule
    \textsc{RAG} & {44.5} & {56.8} & {-} & {-} & {88.1} & {13.8} \\
    \textsc{FiD} & {48.3} & {67.2} & {-} & {-} & {89.5} & {16.9} \\
    \textsc{Evi.} & {49.8} & {67.8} & {-} & {-} & {89.8} & {17.9} \\
    \midrule
    \multicolumn{7}{c}{\textsc{FilCo, top 1}} \\
    \midrule
    \textsc{FilCo} & {44.7} & {59.0} & {60.0} & {73.8} & \textbf{94.2} & \textbf{68.3} \\
    \midrule
    \multicolumn{7}{c}{\textsc{FilCo, top 5}} \\
    \midrule
    \textsc{Full} & {47.6} & {67.3} & {61.5} & {72.7} & {88.0} & {64.8} \\
    \textsc{Psgs} & {52.9} & {69.1} & {62.3} & {73.7} & {90.7} & {64.6} \\
    \textsc{FilCo} & \textbf{61.8} & \textbf{71.1} & \textbf{65.0} & \textbf{73.9} & {91.4} & {66.0} \\
    \textsc{Silver} & {62.0} & {71.1} & {65.2} & {73.9} & {92.2} & {66.1} \\
    \bottomrule
    \end{tabular}
}
\caption{Generation results when providing top-5 retrieved passages filtered by passages or sentences. \textsc{RAG}, \textsc{FiD}, and \textsc{Evi.} are top-performing methods. We \textbf{bold-type} the best results that do not use silver contexts.}
\label{tab:top5-granularity}
\end{table}

\section{Related Work}

\paragraph{Augmented Generation}
Providing additional contexts to generation has shown to be effective~\cite{lewis2020retrieval,guu2020realm,mialon2023augmented} across many knowledge-intensive tasks~\cite{petroni-etal-2021-kilt}. While the most common approach with a set of retrieved passages is to append them all to the input, some works explored the optimal granularity and strategy to do this.
% optimal length
\citet{wang-etal-2019-multi} identify 100 words to be the optimal size for candidate passages, which then became the \textit{de facto} length. Many works explored retrieval at varied granularity, including paragraph \citep{lee-etal-2019-latent,feldman-el-yaniv-2019-multi}, phrase \citep{lee-etal-2021-phrase}, and even token levels \citep{khandelwal2020generalization,alon2022neurosymbolic}, which all reveal a trade-off in difficulty between retrieval and generation: retrieving longer sequences is easier, but it is harder to generate correct output from them. 
In fact, \citet{shi2023large} shows that model performance can dramatically decrease when irrelevant information is included in output-supporting documents. 
Our method alleviates this in-passage distraction, by allowing arbitrary passage sizes at retrieval time, and providing precisely useful content for generation.

\paragraph{Optimizing Retrieval for Augmentation}
Many works focus on post-process retrieved content to augment the generation.
% passage reranking
A common approach is to rerank retrieved passages and provide only the top few under limited input capacity, based on the similarity between query and passages \citep{nogueira2020passage}, the majority of reader predictions \citep{mao-etal-2021-reader}, and utility for generation \citep{wang2018r}.
% \jun{There is a research topic called Answer Sentence Selection (AS2) in the context of question answering. I wonder if we can add one sentence to clarify how our work differs from prior work on the topic.}
% optimal granularity
\citet{asai-etal-2022-evidentiality} measures the evidentiality of retrieved passages to improve context quality, by removing irrelevant passages and skipping the retrieval step \citep{mallen2023trust}. Nonetheless, these methods operate on the coarse passage level, thus still suffering from in-passage distractions. 
Our method has similarities to answer sentence selection \citep{yu2014deep}, which can operate at a more fine-grained sentence level. 
Yet further, our filtering can apply to text split in arbitrary granularity that optimizes the task of interest, and capture more subtle variances in context.

\section{Conclusion and Future Work}

We propose a context filtering method, \textsc{FilCo}, to provide precisely supportive content to assist model generations, which effectively removes distracting content in both passages partially supporting and irrelevant to the queries. Applying our method brings an average of $2.8$ and $3.0$ point increase with \textsc{Flan-T5} and \textsc{Llama2}, across six knowledge-intensive language datasets from question answering, fact verification, to knowledge-grounded dialog generation. 
Our work also reveals varied recipes to effectively filter context for different tasks. We hope that \textsc{FilCo} can facilitate more developments toward faithful generations in more scenarios. 

\section*{Limitations}

Our proposed method has been shown effective across various tasks, however, may be in certain data domains, under automatic evaluation metrics, and with sufficient computational resources. 

Our approach is domain-agnostic in principle, however, all the datasets we experiment with are built from Wikipedia articles, i.e., the open domain. Tasks of other domains such as news \citep{trischler-etal-2017-newsqa}, biomedical knowledge \citep{nentidis2023bioasq}, and even fictional stories \citep{kocisky-etal-2018-narrativeqa,xu-etal-2022-fantastic}, can readily adopt our method and potentially benefit from it. Nonetheless, we encourage readers to verify its effectiveness before directly extrapolating our conclusion to special-domain datasets. 

We evaluate model retrieval, filtering, and generation performance using automatic metrics such as Exact Match and Unigram F1, which have become the standard metrics. Beyond lexical-based metrics, we keep open to neural- or human-based evaluations, given the potentially inaccurate automatic measures, especially with increasingly complex tasks \citep{pugaliya-etal-2019-bend} and models of greater capacities \citep{kamalloo2023evaluating}.

Our method requires training models to (i) filter context, and (ii) generate output, which necessitates certain computational resources, according to the model architecture and size of choice. Nonetheless, our method costs less computation compared to traditional full-passage augmentation. As shown by \S\ref{sec:filter-ablation}, a generation model with filtered content requires at least $4.7$ times less computation, at both training and inference time.

\section*{Acknowledgements}

This work was supported in part by a grant from Bosch.
We thank the members of CMU LTI for their helpful discussion and feedback on this work.

% Entries for the entire Anthology, followed by custom entries
\bibliography{anthology,custom}
\bibliographystyle{acl_natbib}

% \appendix

\end{document}